\documentclass[10pt,conference]{IEEEtran}
\IEEEoverridecommandlockouts
\usepackage[caption=false,font=footnotesize]{subfig}
\usepackage{siunitx}
\usepackage{cite}
\usepackage{amsmath,amssymb,amsfonts}
\usepackage{amsthm}
\usepackage{amsfonts}
\usepackage{algorithmic}
\usepackage{graphicx}
\usepackage{textcomp}
\usepackage{xcolor}
\usepackage[colorlinks=false]{hyperref}
\usepackage{comment}
\def\BibTeX{{\rm B\kern-.05em{\sc i\kern-.025em b}\kern-.08em
    T\kern-.1667em\lower.7ex\hbox{E}\kern-.125emX}}
\begin{document}

\title{Deep Neural Networks with Ordinal Loss for Medical Applications\thanks{This work was supported by a research grant from Intuit.}\thanks{\copyright~2026 IEEE. Personal use of this material is permitted. Permission from IEEE must be obtained for all other uses, in any current or future media, including reprinting/republishing this material for advertising or promotional purposes, creating new collective works, for resale or redistribution to servers or lists, or reuse of any copyrighted component of this work in other works.}}

\author{
\IEEEauthorblockN{Tal Dvora}
\IEEEauthorblockA{\textit{Bar-Ilan University} \\
Ramat Gan, Israel \\
taldvora11@gmail.com}
\and
\IEEEauthorblockN{Rotem Haba}
\IEEEauthorblockA{\textit{Bar-Ilan University} \\
Ramat Gan, Israel \\
rotemhaba2910@gmail.com}
\and
\IEEEauthorblockN{Gonen Singer}
\IEEEauthorblockA{\textit{Bar-Ilan University} \\
Ramat Gan, Israel \\
gonen.singer@biu.ac.il}
}

\maketitle
\begin{abstract}
In many prediction problems in medical applications, target labels exhibit an inherent ordinal structure, where class ordering reflects clinically meaningful severity levels. The cost associated with misclassification is often non-uniform and asymmetric, as errors between distant ordinal categories may have substantially more severe consequences than errors between adjacent ones, and overestimating disease severity may have different clinical implications than underestimating it. Traditional loss functions such as multi-class cross-entropy treat all misclassifications equally and fail to incorporate this ordering information. Recent advances in ordinal regression aim to address this limitation by integrating rank-based structures into deep learning models. In this work, we introduce the \textbf{Ordinal Cross-Entropy (OCE)} framework, a general and architecture-independent approach for learning from ordinal data. The proposed method extends the standard cross-entropy formulation to account for misclassification severity through an ordinal cost matrix while preserving the probabilistic interpretation and optimization benefits of the conventional loss. We provide a theoretical analysis of the OCE gradient behavior and show that it yields smoother optimization dynamics and improved ordinal consistency. Experiments on benchmark datasets show that our method achieves lower prediction error costs and better calibration compared to existing state-of-the-art ordinal approaches, establishing OCE as a simple yet effective solution for ordinal regression in deep neural networks.
\end{abstract}


\section{Introduction}

Deep neural networks (DNNs) have become the leading approach for multi-class classification across numerous domains, including medical image analysis \cite{he2016deep,zhu2017deeplung, shen2017deep, abdou2022literature,chen2025review}, where accurate predictions are often essential for clinical decision-making. The standard framework for multi-class learning relies on the cross-entropy (CE) loss \cite{geng2007automatic} combined with a softmax output layer, which offers stable optimization, probabilistic interpretation, and strong empirical performance. However, CE inherently treats all misclassifications as equally severe and does not incorporate the structure of the output space \cite{bertinetto2020making}.

In many real-world applications, particularly in medicine, the target labels are \textit{ordinal} rather than nominal. Disease stages, severity grades, or risk levels exhibit a natural ordering, and misclassification costs depend not only on the distance between classes but also on the direction of the error. In clinical settings, overestimating disease severity may lead to unnecessary interventions, whereas underestimating severity can result in delayed treatment and substantially worse outcomes. For example, in diabetic retinopathy \cite{gulshan2016development}, cancer staging \cite{abraham2019automated, shen2019deep}, or predict the level of Parkinson’s disease \cite{barbero2021ordinal}, predicting a mild case as severe is less harmful than predicting a severe case as healthy, since the latter combines a larger ordinal distance with a more harmful direction of misclassification, leading to substantially higher clinical risk.

To address this gap, several ordinal learning approaches have been proposed. Methods such as CORAL~\cite{cao2020rank} and CORN \cite{shi2023deep} reformulate the ordinal regression problem into a series of $K\!-\!1$ binary classification subtasks while enforcing rank-monotonicity through shared model weights and ordered bias terms. This design guarantees that the predicted probabilities across the binary tasks are non increasing, ensuring consistent and rank-aware prediction. Unimodal label-regularization techniques~\cite{beckham2017unimodal, liu2020unimodal, da2008unimodal, vargas2022unimodal} encourage predictions centered around the correct class. Vargas et al.~\cite{vargas2022unimodal} introduce a probabilistic beta-based regularization framework that explicitly incorporates ordinal structure into the target labels. Their approach models the ground-truth class as a unimodal distribution generated from a beta density centered on the true ordinal level, effectively smoothing the one-hot target vector in a shape-aware manner that reflects the degree of proximity between classes. As a result, misclassifications to adjacent categories are penalized less severely than errors that span multiple severity levels, aligning the learning objective more closely with the underlying ordinal semantics. The authors apply the beta-regularized loss to both standard softmax classification and a stick-breaking (cumulative probability) formulation \cite{sloane1996introduction}, reporting consistent gains over Poisson-, binomial-, and exponential-based \cite{da2008unimodal,beckham2017unimodal} label smoothing methods across multiple benchmark ordinal image datasets, including diabetic retinopathy grading. Beyond probabilistic label–regularization methods, recent approaches explicitly embed ordinal structure at the architectural level. For instance, OR-DGN \cite{tang2023disease} introduces a regression branch to learn continuous severity scores alongside classification. Similarly, DGN-AGLD \cite{tang2025disease} incorporates a dedicated 'variance predictor' module to model label distributions as asymmetric Gaussians, aiming to capture the direction of disease progression. While these hybrid strategies effectively leverage ordinal relationships, they inherently increase model complexity, require designing specialized network branches, and necessitate balancing multiple loss functions (e.g., cross-entropy combined with distribution or regression losses). In contrast, our proposed Ordinal Cross-Entropy (OCE) offers a purely loss-based solution. By integrating an ordinal cost matrix directly into the standard cross-entropy formulation, OCE enforces distance-dependent penalties without altering the network architecture, preserving the desirable optimization properties of the original loss. Other approaches, such as weighted cross-entropy (WCE)~\cite{aurelio2019learning,akil2020fully}, attempt to incorporate misclassification costs at the class level. Although effective in specific settings, these methods do not fully capture the structure of ordinal misclassification risks, and in particular, they cannot express scenarios in which the \textit{direction} of the error matters. A notable attempt to integrate ordinal distance is the Ordinal Loss (OL) proposed by Chen et al.~\cite{chen2019fully}, in which penalties are determined by the ordinal distance between predicted and true classes. Developed for the specific challenge of grading knee osteoarthritis (OA) severity, the major cause of activity limitation and physical disability in older individuals \cite{conaghan2015impact,neogi2013epidemiology,ortman2014aging,zhao2025value}, this approach addresses a critical limitation of the standard cross-entropy loss, which treats all misclassifications equally regardless of the semantic gap between categories. The authors argue that in clinical ordinal tasks, a ``faraway'' error (e.g., predicting severe Grade~4 when the truth is healthy Grade~0) is significantly more detrimental than a ``nearby'' error (e.g., confusing Grade~0 with Grade~1). To mitigate this, the method utilizes an ``adjustable ordinal matrix'' to define penalty weights that increase as the distance between the predicted rank and the ground truth grows. This mechanism constrains the model's probability output, ensuring that the likelihood assigned to incorrect classes decreases as they become further removed from the true grade.
However, the ordinal loss function does not retain the properties of the cross-entropy; for example, it assumes a linear relationship between the predicted probability and the loss value, instead of the logarithmic relationship in the cross-entropy loss function.

These limitations highlight the need for a loss function that is able to:
\begin{enumerate}
    \item Preserve the desirable optimization properties of cross-entropy.
    \item Incorporate ordinal relationships and distance-dependent penalties.
    \item Encode asymmetric misclassification costs, where the risk of underestimation and overestimation are not equivalent.
\end{enumerate}

To address this need, we propose the \textbf{Ordinal Cross-Entropy (OCE)} framework, a principled extension of cross-entropy for ordinal classification. OCE integrates a general cost matrix allowing domain experts to specify both distance-dependent and \textit{direction-dependent} misclassification penalties, while maintaining the logarithmic sensitivity of CE. This enables the model to prioritize avoidance of clinically dangerous misdiagnoses and learn decision boundaries aligned with real-world risk. As we show through theoretical and empirical evaluation, OCE provides smoother gradient behavior, improved ordinal consistency, and superior performance compared to existing ordinal and cost-sensitive loss formulations.
\\All experiments presented in this paper can be reproduced using the source code available at \url{https://github.com/Taldvora1/Deep-Neural-Networks-with-Ordinal-Loss-for-Medical-Applications}.

\section{Proposed Method: Ordinal Cross-Entropy (OCE)}
The use of a cost matrix to encode unequal misclassification penalties
is well-established in the cost-sensitive learning literature
\cite{elkan2001foundations}, where the learning objective is explicitly
aligned with the real-world consequences of prediction errors.
In this section, we introduce the \textit{Ordinal Cross-Entropy (OCE)} loss, a 
distance-aware and cost-sensitive extension of the standard cross-entropy loss.
OCE is designed to preserve the probabilistic and optimization benefits of
cross-entropy while incorporating both ordinal structure and clinically relevant
misclassification costs.
The use of a cost matrix to encode unequal misclassification penalties
is well-established in the cost-sensitive learning literature
\cite{elkan2001foundations}, where the learning objective is explicitly
aligned with the real-world consequences of prediction errors.
In this section, we introduce the \textit{Ordinal Cross-Entropy (OCE)} loss, a 
distance-aware and cost-sensitive extension of the standard cross-entropy loss.
OCE is designed to preserve the probabilistic and optimization benefits of
cross-entropy while incorporating both ordinal structure and clinically relevant
misclassification costs.

\subsection{Penalty Matrix and Cost Matrix}

We define a square cost matrix $C \in \mathbb{R}^{I \times I}$ with non-negative
entries $c_{i,j} \ge 0$, where $c_{i,j}$ represents the real-world cost incurred
when a sample with true class $y = v_i$ is predicted as $\hat{y} = v_j$. Here,
$\{v_1,\dots,v_I\}$ denotes the ordered set of ordinal class values (e.g., healthy,
mild, moderate, severe), while the indices $i$ and $j$ represent their respective
ordinal positions. To model realistic
ordinal risk behaviour, the cost increases with the ordinal distance
$d = |i - j|$ and may reflect different severities for underestimation and
overestimation. Formally:

\[
C_{i,j} =
\begin{cases}
k_{1} + \lambda_{1} \cdot d, & \text{if } i < j \ (\text{overestimation}), \\
k_{2} + \lambda_{2} \cdot d, & \text{if } i > j \ (\text{underestimation}), \\
0, & \text{if } i = j,
\end{cases}
\]
where $k_1, k_2$ represent baseline costs and $\lambda_1, \lambda_2$ control the
growth rate as classes become further apart. In clinical scenarios where 
underestimation carries greater risk (e.g., diagnosing cancer at a more advanced
stage than reality), one may impose:
\[
k_1 \leq k_2, \qquad \lambda_1 \leq \lambda_2,
\]
while equality recovers the symmetric case.

\vspace{0.1cm}
\noindent\textbf{Transition from Cost Matrix to Penalty Matrix.}
The cost matrix $C$ captures real-world misclassification costs reflecting clinical risk and domain-specific severity. However, these costs are not directly suitable for use within a probabilistic loss function, as they are not normalized and do not explicitly distinguish between penalties for incorrect predictions and benefits for correct ones. We therefore transform $C$ into a penalty matrix $P$ that can be consistently integrated into the OCE loss.

\begin{figure}[t]
\centering

\subfloat[Symmetric penalty structure]{%
    \includegraphics[width=0.85\columnwidth]{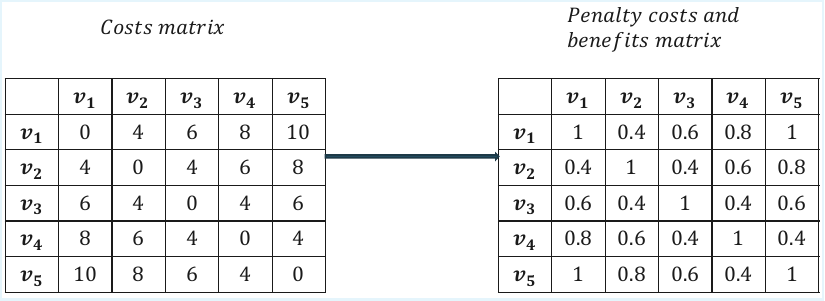}%
    \label{fig:penalty_sym}}
\vskip 0.25cm

\subfloat[Asymmetric penalty structure]{%
    \includegraphics[width=0.85\columnwidth]{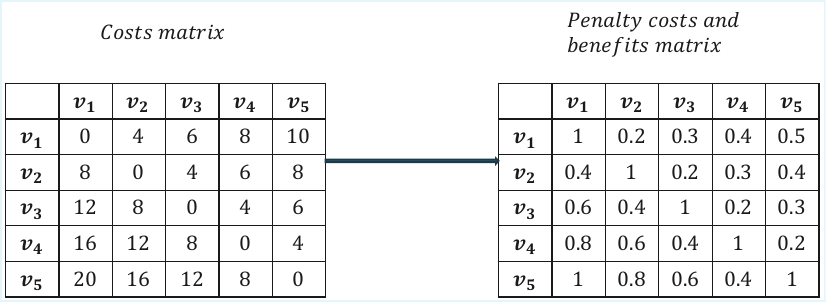}%
    \label{fig:penalty_asym}}

\caption{Comparison of penalty structures for ordinal misclassification.
(a) Symmetric penalties assign equal costs to underestimation and overestimation based on ordinal distance.
(b) Asymmetric penalties reflect domain-specific risk.
In both cases, a maximum diagonal benefit reinforces correct predictions.
}
\label{fig:fig1}
\end{figure}

Accordingly, we construct the penalty matrix $P$
by modifying the diagonal and normalizing the scale as follows:
\[
P_{i,i} = \max(C), \qquad \forall i,
\]

and the matrix is normalized to $[0,1]$ by:
\[
P = \frac{P}{\max(C)}.
\]
This transformation preserves the relative misclassification relationships
defined by the real-world costs while providing stable optimization dynamics and
a clear benefit signal for confident correct predictions. From this point onward, for notational simplicity, we denote the entries of the resulting penalty--reward matrix by $c_{ij}$; diagonal elements $c_{ii}$ represent rewards for correct classification, whereas off-diagonal elements $c_{ij}$ for $i\neq j$ represent misclassification penalties.
Figure~\ref{fig:fig1} illustrates both the real-world cost matrix and the derived
penalty matrix, including the transition process in the symmetric and asymmetric
configurations, for better understanding.

\subsection{Ordinal cross-entropy loss}
The ordinal cross-entropy is a logarithmic function that can be easily optimized using gradient descent methods, similarly to the conventional cross-entropy loss function. 

In the training process of a neural network, the \textit{OCE} loss function aims to find the optimal parameters of the network, which minimize \textit{OCE}. In order to solve the optimization problem using a gradient descent algorithm, we derive the partial derivatives of the \textit{OCE} with respect to the outputs of the neural network before the normalization function, $z_{k,i}, \ \forall k \in \{1,\dots,K\}, \ i \in \{1,\dots,I\}$.

Assume a normalization softmax function, $\mathbb{R}^I \rightarrow \mathbb{R}^I$, that maps each vector of a sample $k$, $\mathbf{z}_k \in \mathbb{R}^I$ to a vector $\hat{p}_k \in \mathbb{R}^I$ such that
\begin{equation}
\begin{aligned}
\hat{p}_{k,i}(\mathbf{z}_k)
 &= 
    \frac{e^{z_{k,i}}}{\sum_{j=1}^{I} e^{z_{k,j}}}, 
    \quad i = 1,\dots,I, \\
0 \leq \hat{p}_{k,i} &\leq 1, \quad 
\sum_{i=1}^{I} \hat{p}_{k,i} = 1.
\end{aligned}
\label{eq:softmax}
\end{equation}

Assuming a one-hot encoded target vector $y_k$, where $y_{k,m_k}=1$ and
$y_{k,i}=0$ for all $i \neq m_k$, with $m_k$ denoting the index of the true
ordinal class of sample $k$, the OCE loss becomes:
{\small
\begin{equation}
\begin{aligned}
\text{OCE} = - \sum_{k=1}^{K} \sum_{i=1}^{I} c_{m_k,i} \Bigg[
y_{k,i} \log \left( \frac{e^{z_{k,i}}}{\sum_{j=1}^{I} e^{z_{k,j}}} \right)  \\
+ (1 - y_{k,i}) \log \left( 1 - \frac{e^{z_{k,i}}}{\sum_{j=1}^{I} e^{z_{k,j}}} \right)
\Bigg].
\end{aligned}
\label{eq:oce_softmax}
\end{equation}
}

Under this assumption, the OCE value for a single sample $k$ can be written as:
\begin{equation}
\begin{aligned}
\text{OCE}_k = - \Big[
& c_{m_k,m_k} \, z_{k,m_k}
- c_{m_k,m_k} \log \left( \sum_{j=1}^{I} e^{z_{k,j}} \right) \\
& + \sum_{\substack{i=1 \\ i \ne m_k}}^{I} c_{m_k,i}
    \Big(
        \log \left( \sum_{j=1}^{I} e^{z_{k,j}} - e^{z_{k,i}} \right)\\
        &- \log \left( \sum_{j=1}^{I} e^{z_{k,j}} \right)
    \Big)
\Big].
\label{eq:oce_sample}
\end{aligned}
\end{equation}

Calculating the derivative for $z_{k,v}$, we get:
\begin{equation}
\begin{aligned}
\nabla_{z_{k,v}} OCE_k =\;
& -\, c_{m_k,m_k}\, \delta_{v,m_k} 
+ c_{m_k,m_k}\, \hat{p}_{k,v} \\
& - \hat{p}_{k,v} \sum_{i \notin \{m_k, v\}}
    \frac{c_{m_k,i}}{1 - \hat{p}_{k,i}}
+ \hat{p}_{k,v} \sum_{i \ne m_k} c_{m_k,i}
\label{eq:oce_grad}
\end{aligned}
\end{equation}

where $\delta_{v,m_k}$ denotes the Kronecker delta, defined as
\[
\delta_{v,m_k} =
\begin{cases}
1, & \text{if } v = m_k, \\
0, & \text{otherwise}.
\end{cases}
\]

\medskip
Note that the development of Equation~\eqref{eq:oce_grad} is presented in the Appendix.
It can be shown that the derivative of \textit{OCE} in Equation~\eqref{eq:oce_grad} maintains several properties, as follows.

\textbf{Property 1.}
The gradient for the observed class $\nabla_{z_{k,m_k}} OCE_k$ is non-positive and
approaches zero as $\hat{p}_{k,m_k}$ increases and the probabilities of unobserved classes decrease.
\begin{proof}
For $v = m_k$:
\begin{equation}
\begin{aligned}
\nabla_{z_{k,m_k}} OCE_k =\,&
-c_{m_k,m_k}(1 - \hat{p}_{k,m_k})\\
&- \hat{p}_{k,m_k}
\sum_{i \ne m_k}
\left(
\frac{c_{m_k,i}}{1 - \hat{p}_{k,i}} - c_{m_k,i}
\right)
\end{aligned}
\label{eq:oce_grad2}
\end{equation}

Since $0 \leq \hat{p}_{k,i} \leq 1$, $\nabla_{z_{k,m_k}} OCE_k$ is non-positive. It can be seen from the first term, $-c_{m_k,m_k}(1 - \hat{p}_{k,m_k})$, that 
$\nabla_{z_{k,m_k}} OCE_k$ decreases in magnitude as $\hat{p}_{k,m_k}$ increases and the remaining probabilities decrease.
From the second term, $\hat{p}_{k,m_k} \sum_{i \ne m_k} \left( \frac{c_{m_k,i}}{1 - \hat{p}_{k,i}} - c_{m_k,i} \right)$, we see that
$\nabla_{z_{k,m_k}} OCE_k$ further decreases in magnitude as $\hat{p}_{k,i}, \ \forall i \ne m_k,$ decreases.
Furthermore, a change in predicted probabilities of unobserved classes with high costs $c_{m_k,i}$ (high risk)
has a greater effect than those with low penalty costs.
\end{proof}
\textbf{Property 2.} The gradient for an unobserved class $\nabla_{z_{k,v}} OCE_k$, $\forall v \ne m_k$, is proportional to  $\hat{p}_{k,v}$ and is non-negative whenever $c_{m_k,m_k} + \sum_{i \ne m} c_{m_k,i} \geq \sum_{i \ne m,v} \frac{c_{m_k,i}}{1 - \hat{p}_{k,i}}$. In this case, the gradient increases in magnitude as $\hat{p}_{k,v}$ increases.
\begin{proof}
For an unobserved class $v \ne m_k$, the gradient of the OCE loss with respect to
$z_{k,v}$ can be written as
\[
\nabla_{z_{k,v}} OCE_k
=
\hat{p}_{k,v}
\left(
c_{m_k,m_k}
+ \sum_{i \ne m_k} c_{m_k,i}
- \sum_{\substack{i \ne m_k \\ i \ne v}}
\frac{c_{m_k,i}}{1 - \hat{p}_{k,i}}
\right).
\]

Since $\hat{p}_{k,v} \ge 0$, the sign of the gradient is determined by the
bracketed term. Hence, $\nabla_{z_{k,v}} OCE_k$ is non-negative whenever the
bracketed expression is non-negative. Under this condition, the magnitude of the
gradient increases as $\hat{p}_{k,v}$ increases, as it appears as a multiplicative
factor.
\end{proof}

\textbf{Property 3.} The gradient satisfies $\nabla_{z_{k,v}} OCE_k = 0$ for all $v$ under perfect classification, i.e., when $\hat{p}_{k,m_k}=1$ and $\hat{p}_{k,v}=0$ for all $v \neq m_k$.

\begin{proof}
From Property~1, the gradient for the observed class satisfies
$\nabla_{z_{k,m_k}} OCE_k = 0$ when $\hat{p}_{k,m_k}=1$ and
$\hat{p}_{k,v}=0$ for all $v \neq m_k$.
From Property~2, the gradient for any unobserved class $v \neq m_k$ is
proportional to $\hat{p}_{k,v}$ and therefore equals zero when
$\hat{p}_{k,v}=0$.
Hence, $\nabla_{z_{k,v}} OCE_k = 0$ for all $v$ under perfect classification.
\end{proof}
It can be seen that when the benefit associated with correct classification
dominates the penalties of misclassification, namely,
\[
c_{m_k,m_k} \gg c_{m_k,i}, \quad \forall i \neq m_k,
\]
the condition
\[
c_{m_k,m_k}
+ \sum_{i \neq m_k} c_{m_k,i}
\;\ge\;
\sum_{\substack{i \neq m_k \\ i \neq v}}
\frac{c_{m_k,i}}{1 - \hat{p}_{k,i}}
\]

is satisfied. Under this condition, the gradient
\[
\nabla_{z_{k,v}} OCE_k, \quad v \neq m_k,
\]
is non-negative and increases in magnitude as $\hat{p}_{k,v}$ increases.

However, when there exists an unobserved class $i \neq m_k, v$ such that
$\hat{p}_{k,i} \to 1$, the above condition no longer holds, and instead
\[
c_{m_k,m_k}
+ \sum_{i \neq m_k} c_{m_k,i}
\;<\;
\sum_{\substack{i \neq m_k \\ i \neq v}}
\frac{c_{m_k,i}}{1 - \hat{p}_{k,i}}.
\]

\begin{figure}[t]
\centering
\includegraphics[width=\linewidth]{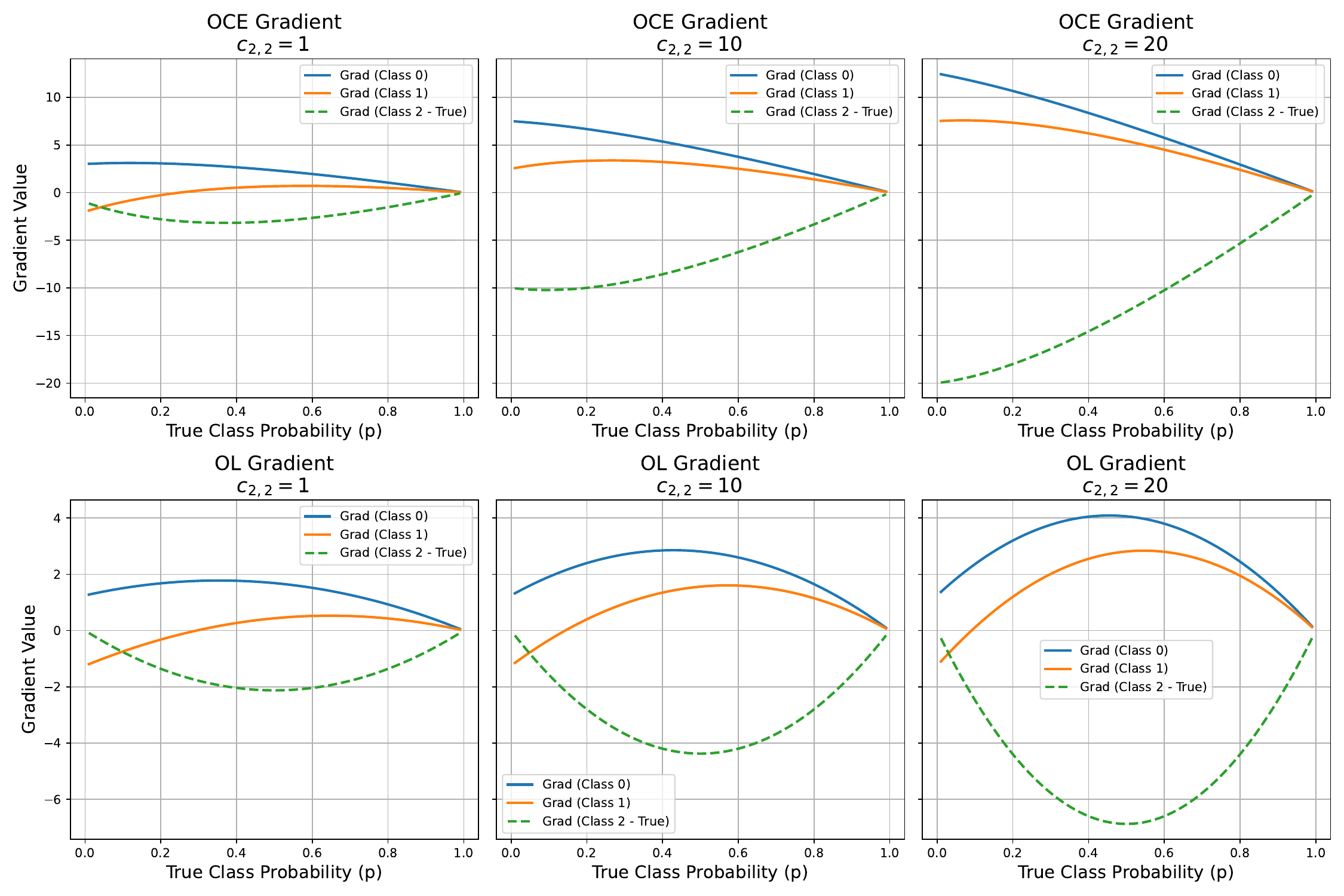}
\caption{
Visualization of the gradient dynamics for Ordinal Cross-Entropy (OCE) 
and Ordinal Loss (OL) across different values of the true-class penalty 
$c_{2,2}$.
}
\label{fig:fig2}
\end{figure}

In this case, the bracketed term in the gradient expression becomes negative,
and consequently $\nabla_{z_{k,v}} OCE_k < 0$ for that class $v \neq m_k$. In other words, increasing $z_{k,v}$, or equivalently the predicted probability
$\hat{p}_{k,v}$, at the expense of a highly confident unobserved class
$\hat{p}_{k,i}$ may decrease the OCE value. Figure~\ref{fig:fig2} shows the OCE gradient behavior compared to OL gradient behavior. 

\subsection{Intuition and Interpretation}

The OCE loss preserves the logarithmic sensitivity of cross-entropy while
modifying it to account for the severity of ordinal errors. For the true class
$m_k$, the reward for correct prediction is weighted by the corresponding
benefit $c_{m_k,m_k}$, whereas for incorrect classes the penalty increases as a
function of both the model's confidence $\hat{p}_{k,i}$ and the associated
misclassification cost $c_{m_k,i}$. Consequently, high-probability and
clinically severe mistakes contribute substantially more to the loss than
low-risk or near-adjacent errors.

\begin{figure}[t] \centering \includegraphics[width=\columnwidth]{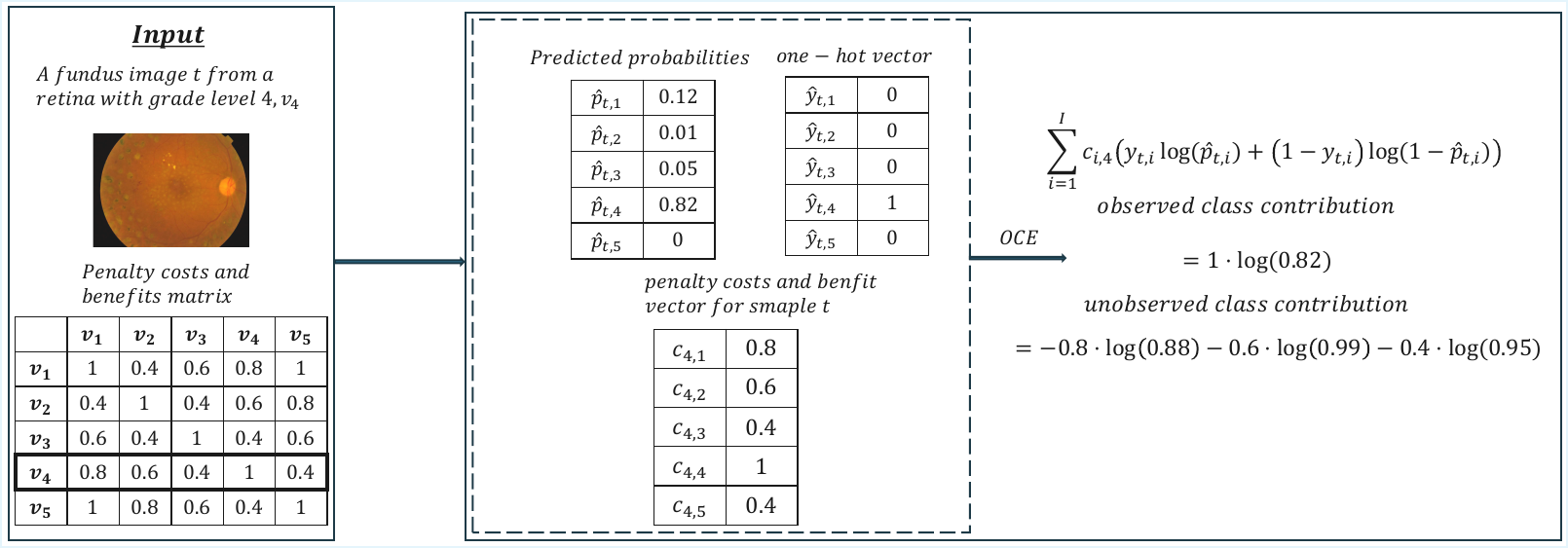} \caption{ Illustration of the proposed ordinal cross-entropy loss for a diabetic
retinopathy sample with true class 4, using the penalty matrix from ~\ref{fig:penalty_sym}. } \label{fig:fig3} \end{figure}

Figure~\ref{fig:fig3} illustrates the OCE computation for a retina diagnosed
with grade level 4, using the symmetric ordinal penalty matrix presented in
\figurename~\ref{fig:penalty_sym}. In this example, correct predictions are rewarded
with a benefit value of $10$, emphasizing the importance of confident
predictions when distinguishing between late-stage and early-stage diagnoses.

\begin{figure}[t]
\centering
\includegraphics[width=\linewidth]{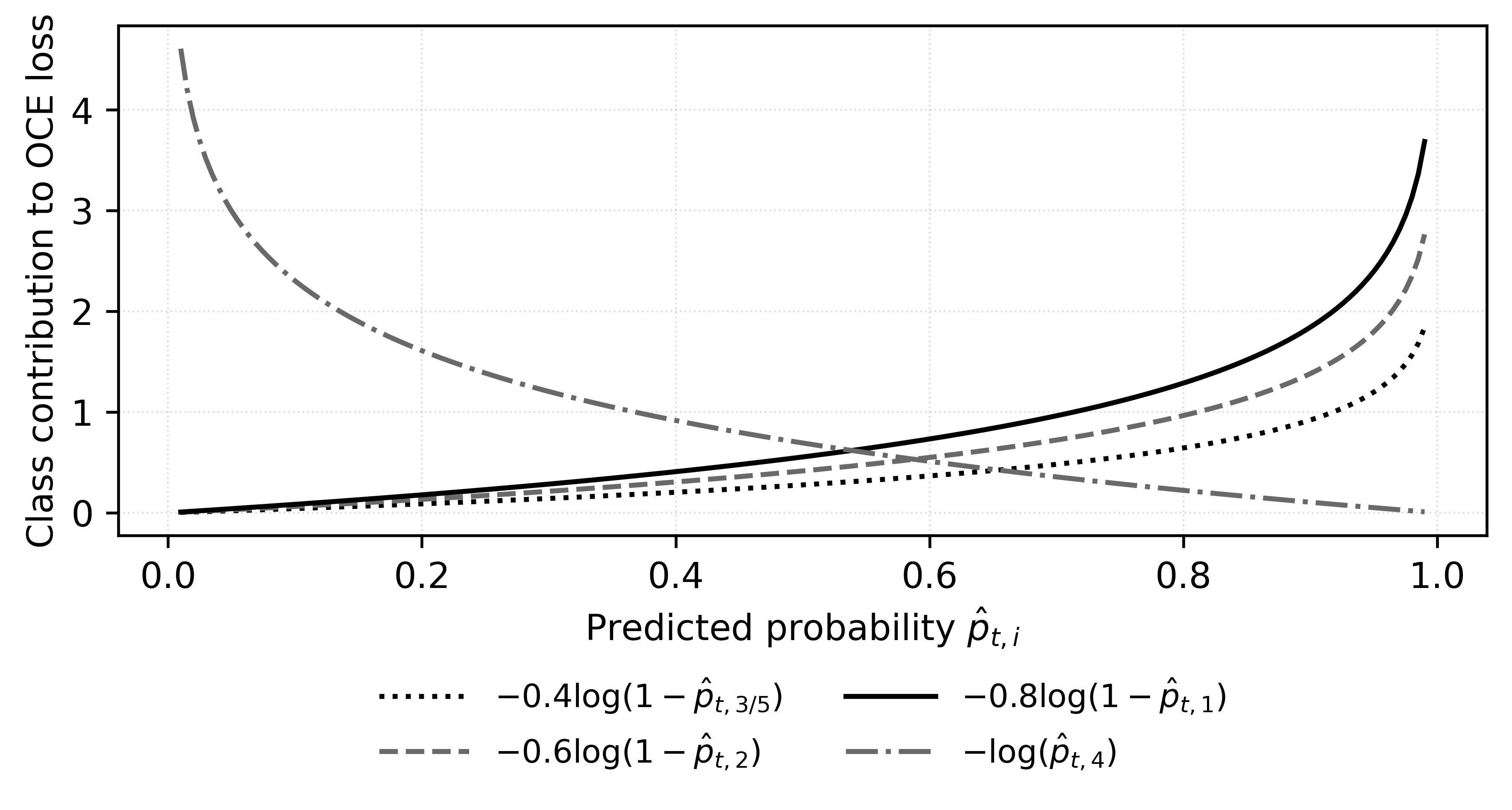}
\caption{
Contribution of class $i$ to the ordinal cross-entropy for a sample $t$
with observed class $v_4$, shown for a range of predictive probabilities
$\hat{p}_{t,:}$ based on the penalty vector $c_{:,4}$ from Figure~\ref{fig:fig3}.
}
\label{fig:fig4}
\end{figure}

Figure~\ref{fig:fig4} illustrates how the contributions of individual classes to
the OCE loss vary as the predicted probability of the true class
$\hat{p}_{t,4}$ increases. As confidence in the correct class grows, its loss
contribution decreases logarithmically, while penalties from unobserved classes
diminish accordingly. Notably, the observed class has a dominant influence on
the loss, and unobserved classes with higher misclassification costs contribute
more strongly, reflecting ordinal distance and task-specific risk.

\section{Experiments and Results}

The primary objective of our experiments is to evaluate whether the proposed
OCE loss effectively \textit{minimizes the total misclassification cost} defined
by a task-specific ordinal cost matrix. Standard metrics such as accuracy, AUC,
and MAE are reported as complementary indicators, but cost minimization is the
central criterion, as it directly reflects the real-world severity of ordinal
errors. To assess the robustness of OCE under different risk profiles, we
conduct experiments using both \textit{symmetric} cost matrices, where
overestimation and underestimation are penalized equally, and
\textit{asymmetric} cost matrices, which assign different costs to
overestimation and underestimation to reflect direction-dependent,
application-specific risk.

\subsection{Dataset}



The APTOS 2019 Blindness Detection dataset consists of retinal fundus images
for automated diabetic retinopathy (DR) grading, released as part of the Kaggle
\href{https://www.kaggle.com/competitions/aptos2019-blindness-detection}
{APTOS 2019 Blindness Detection} competition. Each image is labeled with an ordinal
severity grade from 0 (``No DR'') to 4 (``Proliferative DR'').

The training set includes 3,662 labeled RGB images with varying resolutions and
substantial variability in acquisition conditions, as well as a strong class
imbalance toward the ``No DR'' class, making the task particularly challenging
and well-suited for ordinal and cost-sensitive learning.
\begin{figure}[t]
\centering
\begin{minipage}{0.18\columnwidth}
    \centering
    \includegraphics[
        width=\linewidth,
        height=\linewidth,
        keepaspectratio=false
    ]{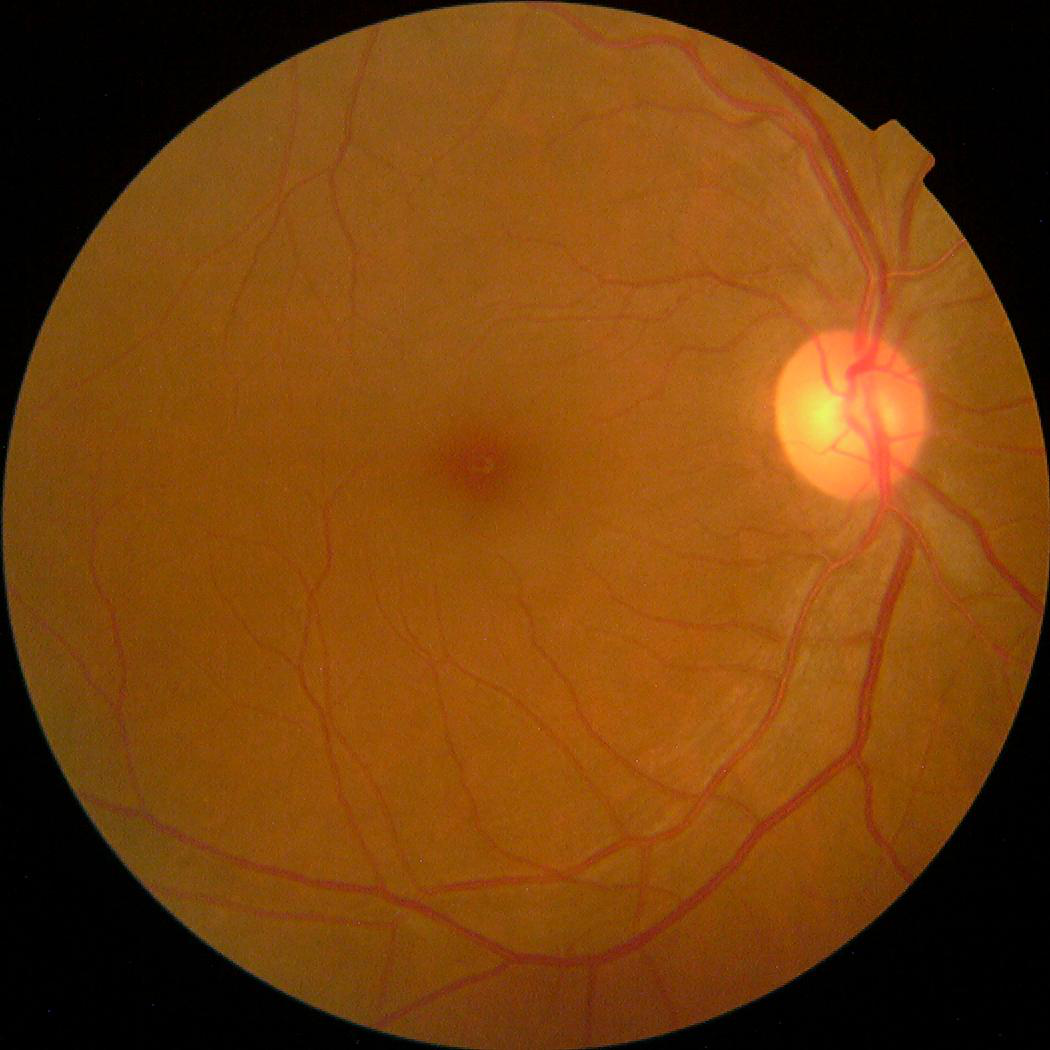}
    \par\vspace{3pt}
    \footnotesize\textbf{Normal}
\end{minipage}\hfill
\begin{minipage}{0.18\columnwidth}
    \centering
    \includegraphics[
        width=\linewidth,
        height=\linewidth,
        keepaspectratio=false
    ]{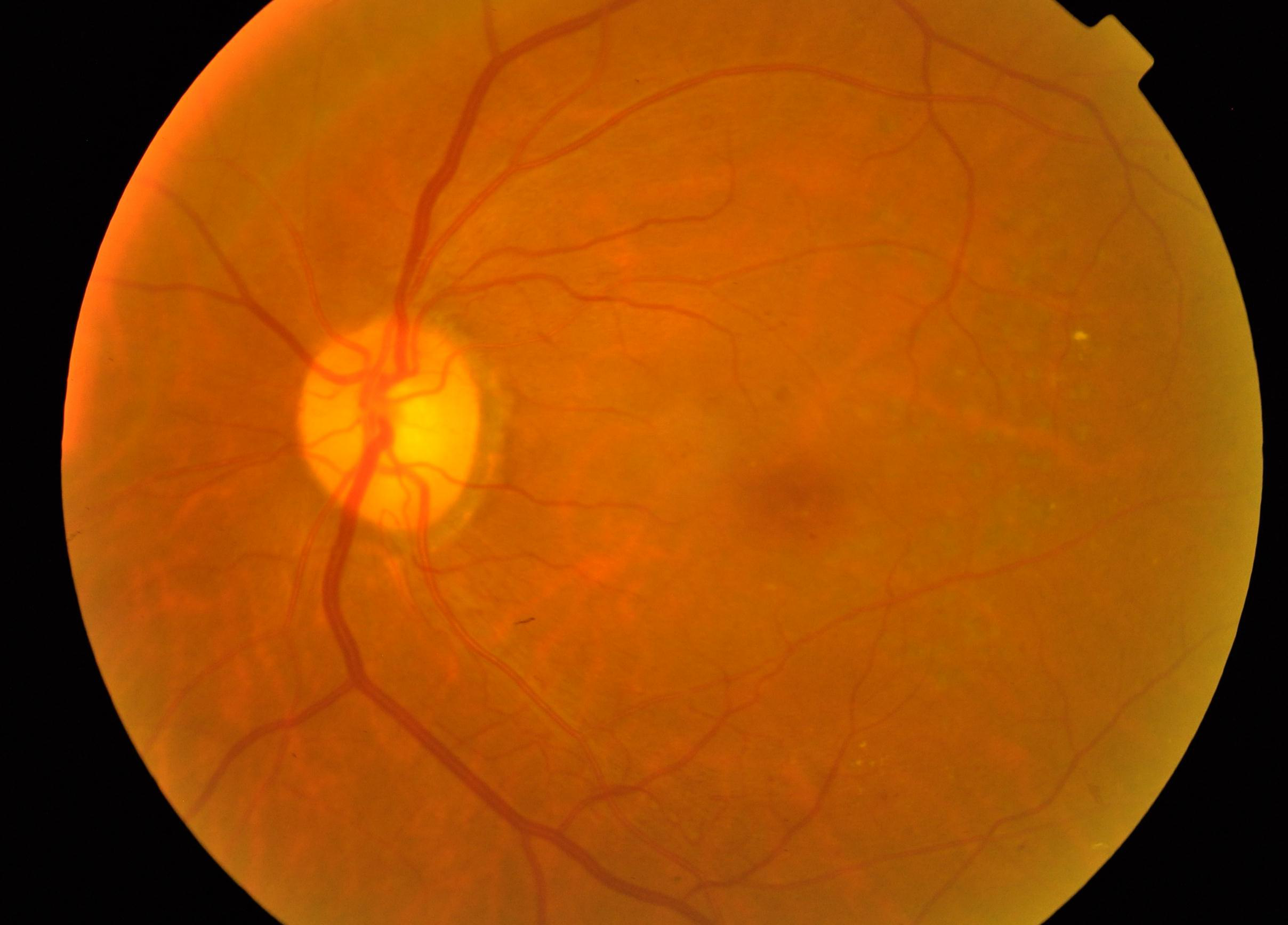}
    \par\vspace{3pt}
    \footnotesize\textbf{Mild}
\end{minipage}\hfill
\begin{minipage}{0.18\columnwidth}
    \centering
    \includegraphics[
        width=\linewidth,
        height=\linewidth,
        keepaspectratio=false
    ]{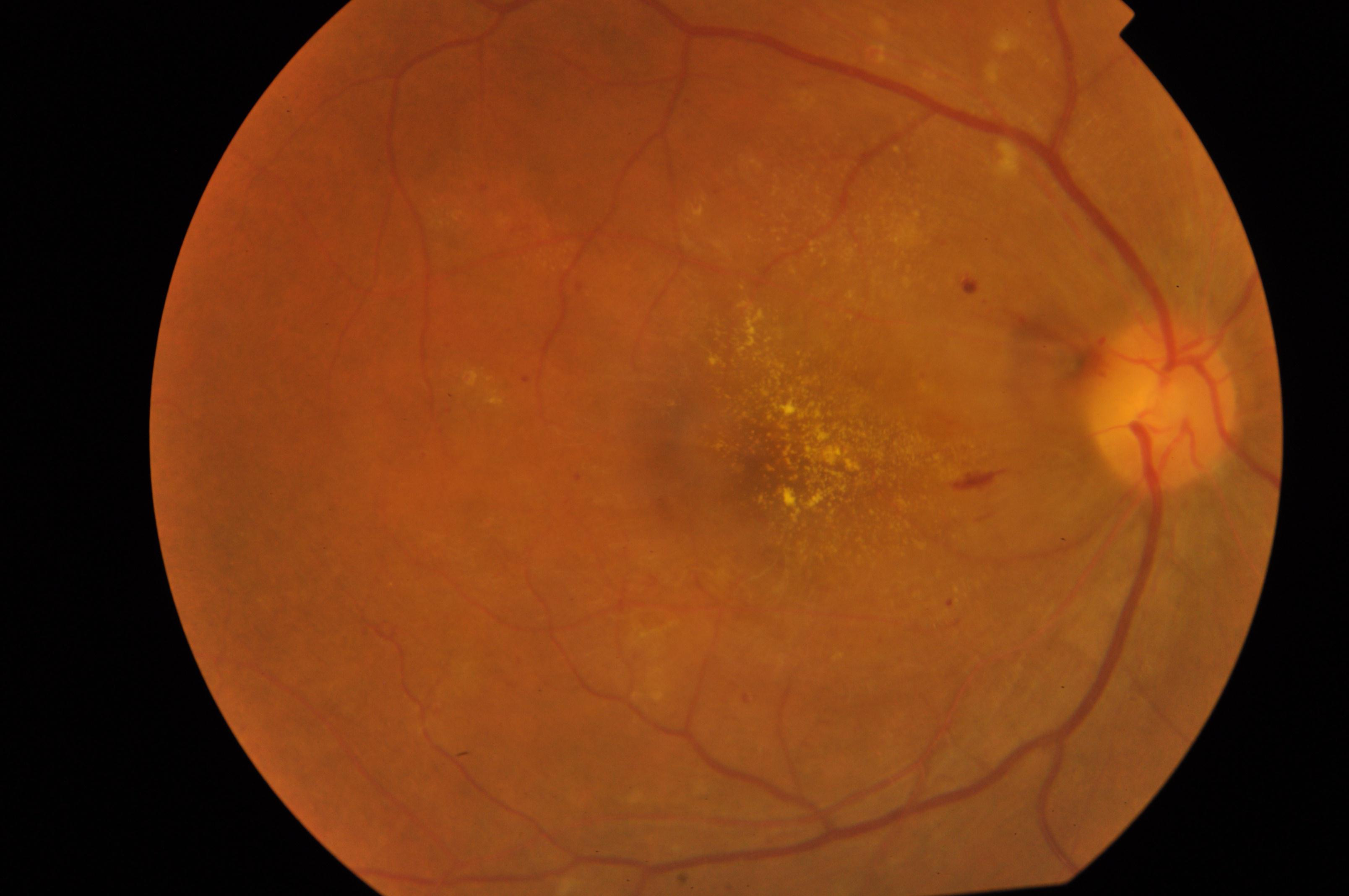}
    \par\vspace{3pt}
    \footnotesize\textbf{Moderate}
\end{minipage}\hfill
\begin{minipage}{0.18\columnwidth}
    \centering
    \includegraphics[
        width=\linewidth,
        height=\linewidth,
        keepaspectratio=false
    ]{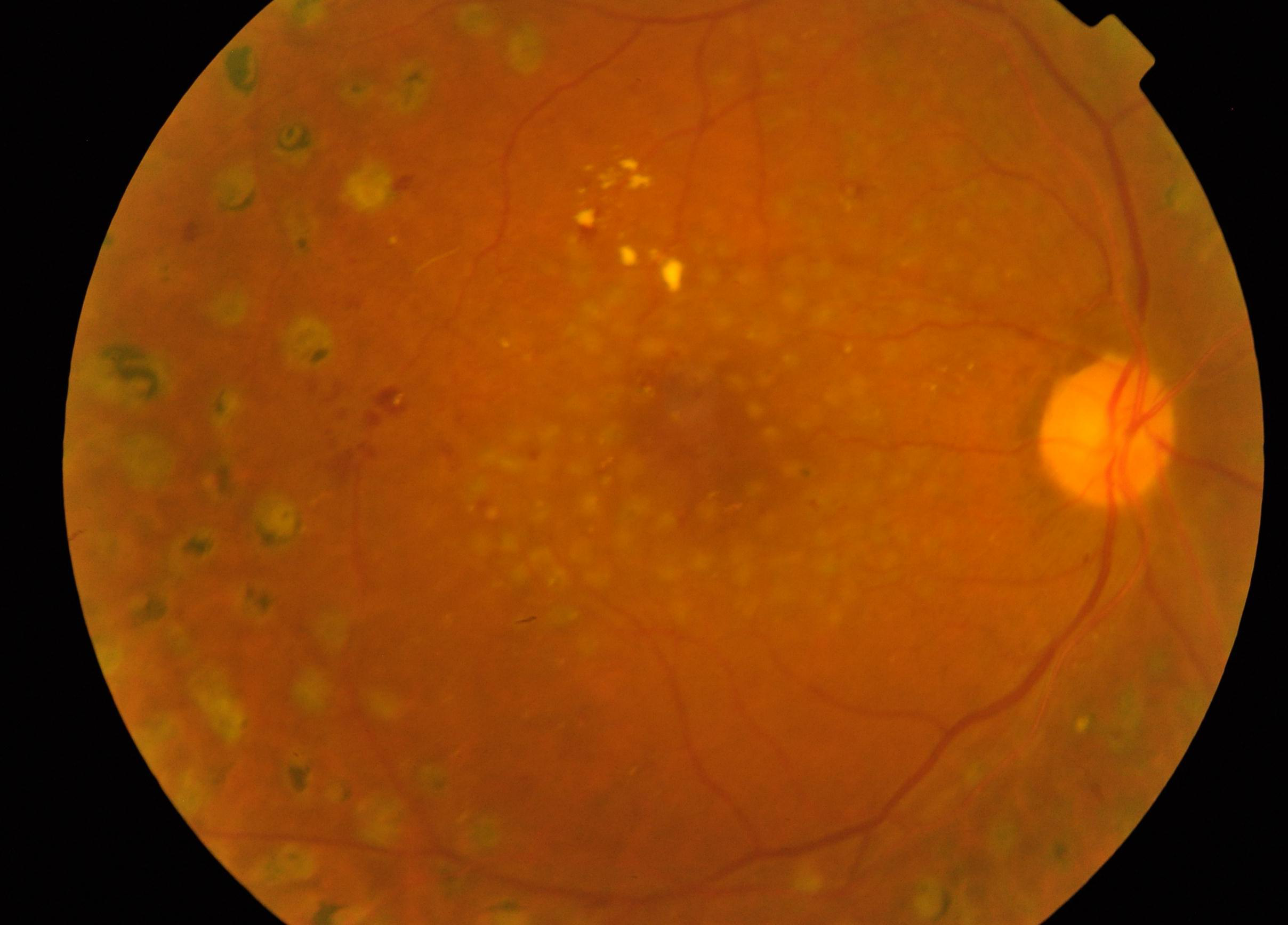}
    \par\vspace{3pt}
    \footnotesize\textbf{Severe}
\end{minipage}\hfill
\begin{minipage}{0.18\columnwidth}
    \centering
    \includegraphics[
        width=\linewidth,
        height=\linewidth,
        keepaspectratio=false
    ]{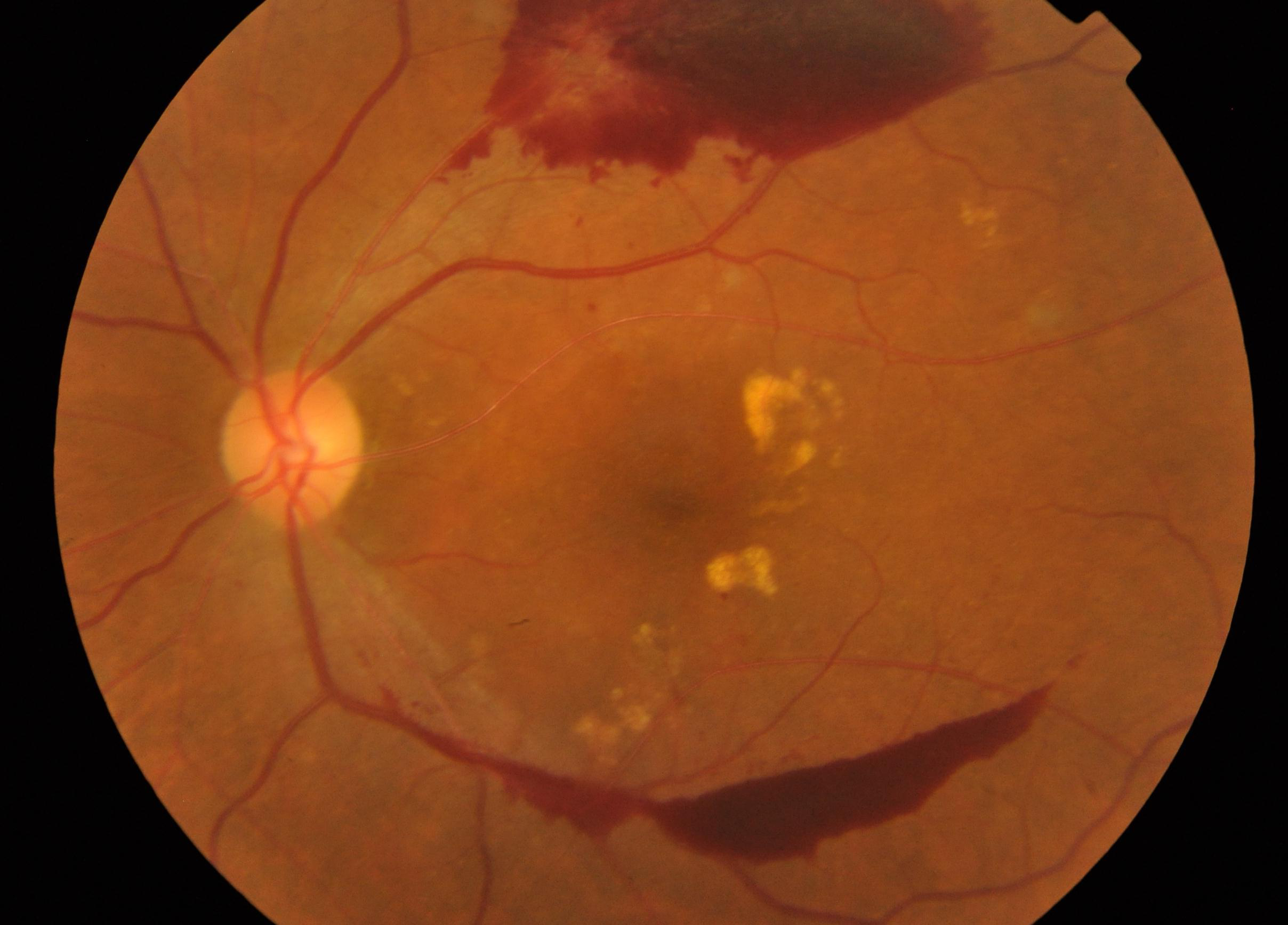}
    \par\vspace{3pt}
    \footnotesize\textbf{Proliferative}
\end{minipage}

\caption{Example images from all five diabetic retinopathy (DR) severity levels.}
\label{fig:dr_examples}
\end{figure}

\begin{table}[!t]
\renewcommand{\arraystretch}{1.3}
\caption{Distribution of DR Grades in the APTOS 2019 Training Set}
\label{tab:dr_distribution}
\centering
\begin{tabular}{ccc}
\hline
\textbf{DR Grade} & \textbf{Grade Name} & \textbf{Total Images} \\ \hline
0 & No DR & 1805 \\
1 & Mild DR & 370 \\
2 & Moderate DR & 999 \\
3 & Severe DR & 193 \\
4 & Proliferative DR & 295 \\ \hline
\end{tabular}
\end{table}

\subsection{Cost Matrix}

Figure~\ref{fig:penalty_sym} illustrates the symmetric ordinal cost structure used for
direct comparison with existing methods, while \figurename~\ref{fig:penalty_asym} presents an
asymmetric variant that assigns different penalties to overestimation and
underestimation, reflecting application-specific risk profiles.

\subsection{Architectures and Experimental Setup}

To assess the robustness of the proposed OCE loss across different network
capacities and feature extraction strategies, we evaluate it on three widely
used convolutional architectures: VGG19, InceptionV3, and DenseNet121. All models are initialized with ImageNet pre-trained weights and trained under
identical experimental conditions to isolate the effect of the loss function.
Training is performed for 25 epochs using the Adam optimizer with a default learning
rate of $1\mathrm{e}{-3}$ and a batch size of 32.

We compare seven loss functions: the proposed Ordinal Cross-Entropy (OCE) with
penalty matrix $P$; standard cross-entropy (CE); cross-entropy with unimodal
beta-based label regularisation (CE-$\beta$)~\cite{vargas2022unimodal}; cross-entropy with poisson label regularisation (CE-P)~\cite{da2008unimodal}; cross-entropy with binomial label regularisation (CE-B)~\cite{da2008unimodal}; cross-entropy with exponential label regularisation (CE-E)~\cite{liu2020unimodal}; and Ordinal Loss (OL) based on distance-aware penalty designs proposed in Chen et al.~\cite{chen2019fully}. For OL, we evaluate several distance-based penalty designs proposed by Chen
et al.~\cite{chen2019fully}, and report results of the design that
achieves the best empirical performance among the evaluated variants.
To evaluate asymmetric ordinal risk, we modify only loss functions that explicitly incorporate a penalty matrix. For OCE, we replace the symmetric matrix with the asymmetric design shown in Figure.~1b. For OL, we adapt the distance-based penalty matrix used in the symmetric setting by halving the
upper triangular entries, thereby imposing milder penalties on overestimation.
For CE and its regularized variants, which do not incorporate penalty matrices into the loss, asymmetry is applied only at the evaluation stage.
\subsection{Evaluation Metrics}

The primary metric for evaluating model performance is the
\textit{Cost}, defined as the average misclassification cost
induced by a predefined cost matrix $C$.

In practice, the cost error is computed as the product of the confusion
matrix and the cost matrix $C$, normalized by the total number of samples:
\[
\text{Cost Error} = \frac{1}{N} \sum_{k=1}^{N} c_{{y}_k, \hat{y_k}},
\]
where $y_k$ and $\hat{y}_k$ denote the true and predicted class labels of
sample $k$, respectively. This metric directly reflects the severity of
ordinal misclassification errors according to their real-world impact.

In addition, we report the following complementary evaluation metrics: Accuracy, Area Under the ROC Curve (AUC), Mean Absolute Error (MAE) and Quadratic Weighted Kappa (QWK) \cite{de2018weighted}. These metrics provide supplementary performance insights; however, they do
not replace the cost-based evaluation, which remains the primary criterion
in this work.

\subsection{Results}

Table~\ref{tab:results} summarizes the average validation performance across
five cross-validation folds for all evaluated architectures under a symmetric
cost matrix. Across all architectures, the proposed OCE loss consistently
achieves the best or second-best misclassification Cost, which is
the primary evaluation criterion. At the same time, OCE maintains competitive
MAE and strong discriminative performance, with accuracy, AUC, and QWK values
comparable to the competing loss functions.

\begin{table}[t]
\centering
\caption{Average validation performance across 5 folds under a symmetric cost matrix. Bold and underlined values indicate the best and second-best results.}
\label{tab:results}
\sisetup{round-mode=places, round-precision=3}
\begin{tabular}{lccccc}
\hline
\textbf{DenseNet121} & Acc & AUC & Cost & MAE & QWK \\ \hline
OCE (ours)     
& \underline{\num{0.7837}} 
& \textbf{\num{0.8965}} 
& \underline{\num{1.0810}} 
& {\num{0.3242}}
& \num{0.8030491566710927}\\
CE-$\beta$ \cite{vargas2022unimodal} 
& \num{0.7817} 
& \num{0.8536} 
& \num{1.0954} 
& \num{0.3294}
&\num{0.804357383303471}\\
CE-P \cite{da2008unimodal} 
& \num{0.756140699} 
& \num{0.860120975} 
& \num{1.219291138} 
& \num{0.365786}
& \num{0.766399}\\
CE-B \cite{da2008unimodal} 
& \num{0.77416812} 
& \num{0.784486095} 
& \num{1.096063964} 
& \underline{\num{0.322200}}
& \underline{\num{0.822974}}\\
CE-E \cite{liu2020unimodal} 
& \textbf{\num{0.788920609}} 
& \num{0.82314662} 
& \textbf{\num{1.04294599}} 
& \textbf{\num{0.3103940}}
& \textbf{\num{0.823619}}\\
CE      
& \num{0.7810} 
& \underline{\num{0.8955}} 
& \num{1.0980} 
& \num{0.3301} 
& \num{0.798212176152261}\\

OL \cite{chen2019fully}      
& \num{0.7401} 
& \num{0.7135} 
& \num{1.2599} 
& \num{0.3700}
& \num{0.7784897463544629}\\ \hline

\textbf{InceptionV3} & Acc & AUC & Cost & MAE & QWK \\ \hline
OCE (ours)     
& \underline{\num{0.7719}} 
& \textbf{\num{0.8783}} 
& \underline{\num{1.1465}} 
& \num{0.3451}
& \num{0.7909391389622376}\\
CE-$\beta$ \cite{vargas2022unimodal} 
& \num{0.7640} 
& \underline{\num{0.8279}} 
& \num{1.1741} 
& \num{0.3510}
& \num{0.7933295856213756}\\
CE-P \cite{da2008unimodal} 
& \num{0.742700223} 
& \num{0.825618898} 
& \num{1.307140672} 
& \num{0.39626352}
& \num{0.7673125176653276}\\
CE-B \cite{da2008unimodal} 
& \num{0.739429583} 
& \num{0.694130634} 
& \num{1.26253763} 
& \num{0.370698132}
& \underline{\num{0.802493663546597}}\\
CE-E \cite{liu2020unimodal} 
& \num{0.762040} 
& \num{0.803051325} 
& \num{1.164220976} 
& \underline{\num{0.344150}}
& \textbf{\num{0.806920}}\\
CE      
& \textbf{\num{0.7738}} 
& \textbf{\num{0.8777}} 
& \textbf{\num{1.1347}} 
& \textbf{\num{0.3412}}
& \num{0.7946371886081327}\\
OL \cite{chen2019fully}       
& \num{0.7335} 
& \num{0.6620} 
& \num{1.3071} 
& \num{0.3871}
& \num{0.7607717948519643}
\\ \hline

\textbf{VGG19} & Acc & AUC & Cost & MAE & QWK \\ \hline
OCE (ours)     
& \underline{\num{0.7581}} 
& \num{0.8214} 
& \textbf{\num{1.2199}} 
& \textbf{\num{0.3681}}
& \textbf{\num{0.7743471146031833}}
\\
CE-$\beta$ \cite{vargas2022unimodal} 
& \textbf{\num{0.7591}} 
& \textbf{\num{0.8589}} 
& \num{1.2678} 
& \num{0.3930}
& \num{0.7196294258620541}\\
CE-P \cite{da2008unimodal} 
& \num{0.754176706} 
& \num{0.762166965} 
& \num{1.266480} 
& \num{0.387416}
& \num{0.729103}\\
CE-B \cite{da2008unimodal} 
& \num{0.749913874} 
& \num{0.806637461} 
& \num{1.292054412} 
& \num{0.395941}
& \num{0.730584}\\
CE-E \cite{liu2020unimodal} 
& \textbf{\num{0.758763113}} 
& \underline{\num{0.84508367}} 
& \num{1.248788602} 
& \num{0.383157}
& \num{0.731722}\\
CE      
& \num{0.7561} 
& \num{0.8269} 
& \underline{\num{1.2442}} 
& \underline{\num{0.3782}}
& \underline{\num{0.7703241095802925}}\\
OL \cite{chen2019fully}       
& \num{0.7260} 
& \num{0.6770} 
& \num{1.3655} 
& \num{0.4087}
& \num{0.7348738002132954}\\
\hline
\end{tabular}
\end{table}

\begin{table}[t]
\centering
\caption{Average validation performance across 5 folds under an asymmetric cost matrix. Bold and underlined values indicate the best and second-best results.}

\label{tab:asy_results}
\sisetup{round-mode=places, round-precision=3}
\begin{tabular}{lccccc}
\hline
\textbf{DenseNet121} & Acc & AUC & Cost & MAE & QWK \\ \hline
OCE (ours)     
& \underline{\num{0.787930026}} 
& \textbf{\num{0.898287965}} 
& \underline{\num{1.754911862}} 
& \underline{\num{0.312356604}}
& \num{0.8199028232715624}\\
CE-$\beta$ \cite{vargas2022unimodal} 
& \num{0.7817} 
& \num{0.8536} 
& \num{1.863651262} 
& \num{0.3294}
&\num{0.804357383303471}\\
CE-P \cite{da2008unimodal} 
& \num{0.756140699} 
& \num{0.860120975} 
& \num{2.113434} 
& \num{0.365786}
& \num{0.766399}\\
CE-B \cite{da2008unimodal} 
& \num{0.77416812} 
& \num{0.784486095} 
& \num{1.811945} 
& \num{0.322200}
& \underline{\num{0.822974}}\\
CE-E \cite{liu2020unimodal} 
& \textbf{\num{0.788920609}} 
& \num{0.82314662} 
& \textbf{\num{1.742407}} 
& \textbf{\num{0.3103940}}
& \textbf{\num{0.823619}}\\
CE      
& \num{0.7810} 
& \underline{\num{0.8955}} 
& \num{1.883316945} 
& \num{0.3301} 
& \num{0.798212176152261}\\
OL \cite{chen2019fully}       
& \num{0.738772236} 
& \num{0.661911787} 
& \num{2.111452872} 
& \num{0.373975746}
& \num{0.7732447230158438}\\
\hline

\textbf{InceptionV3} & Acc & AUC & Cost & MAE & QWK \\ \hline
OCE (ours)     
& \textbf{\num{0.773838105}} 
& \textbf{\num{0.879728841}} 
& \textbf{\num{1.859757989}} 
& \textbf{\num{0.339888561}}
&\num{0.7995625053673364}\\
CE-$\beta$ \cite{vargas2022unimodal} 
& \underline{\num{0.7640}} 
& \num{0.8279} 
& \num{1.972468043} 
& \num{0.3510}
& \num{0.7933295856213756}\\
CE-P \cite{da2008unimodal} 
& \num{0.742700223} 
& \num{0.825618898} 
& \num{2.300924} 
& \num{0.39626352}
& \num{0.7673125176653276}\\
CE-B \cite{da2008unimodal} 
& \num{0.739429583} 
& \num{0.694130634} 
& \num{2.030831} 
& \num{0.370698132}
& \underline{\num{0.802493663546597}}\\
CE-E \cite{liu2020unimodal} 
& \num{0.762040} 
& \num{0.803051325} 
& \num{1.917436} 
& \num{0.344150}
& \textbf{\num{0.806920}}\\
CE      
& \textbf{\num{0.7738}} 
& \underline{\num{0.8777}} 
& \underline{\num{1.900360538}}
& \underline{\num{0.3412}}
&\num{0.7946371886081327}\\
OL \cite{chen2019fully}       
& \num{0.694512087} 
& \num{0.644087227} 
& \num{2.622921843} 
& \num{0.446738774}
& \num{0.7027636201946625}
\\
\hline

\textbf{VGG19} & Acc & AUC & Cost & MAE & QWK \\ \hline
OCE (ours)     
& \num{0.754832} 
& \underline{\num{0.851181}} 
& \textbf{\num{2.054425}} 
& \underline{\num{0.380858735}}
& \underline{\num{0.7673125176653276}}\\
CE-$\beta$ \cite{vargas2022unimodal} 
& \textbf{\num{0.7591}} 
& \textbf{\num{0.8589}} 
& \num{2.288430023} 
& \num{0.3930}
& \num{0.7196294258620541}\\
CE-P \cite{da2008unimodal} 
& \num{0.754176706} 
& \num{0.762166965} 
& \num{2.257635} 
& \num{0.387416}
& \num{0.729103}\\
CE-B \cite{da2008unimodal} 
& \num{0.749913874} 
& \num{0.806637461} 
& \num{2.306163} 
& \num{0.395941}
& \num{0.730584}\\
CE-E \cite{liu2020unimodal} 
& \textbf{\num{0.758763113}} 
& \num{0.84508367} 
& \num{2.245850} 
& \num{0.383157}
& \num{0.731722}\\
CE      
& \underline{\num{0.7561}} 
& \num{0.8269} 
& \underline{\num{2.068174369}} 
& \textbf{\num{0.3782}} 
& \textbf{\num{0.7703241095802925}}\\
OL \cite{chen2019fully}       
& \num{0.648283652} 
& \num{0.669932145} 
& \num{3.517968394} 
& \num{0.602425434}
&\num{0.5216150305067406}\\
\hline
\end{tabular}
\end{table}

Table~\ref{tab:asy_results} reports the average validation performance across
five cross-validation folds for all evaluated architectures under an
asymmetric cost matrix. Across all architectures, the proposed OCE loss
consistently achieves the best or second-best performance in terms of
misclassification Cost and MAE, which are the primary criteria under
asymmetric risk settings.

In addition, OCE maintains competitive and often superior performance in
complementary metrics, including accuracy, AUC, and QWK, indicating that
explicitly modeling asymmetric ordinal risk does not compromise overall
discriminative ability. These results demonstrate the robustness of OCE to
asymmetric cost structures and its effectiveness in prioritizing clinically
meaningful errors.

Overall, the asymmetric setting amplifies the advantage of OCE in minimizing
risk-weighted ordinal errors, demonstrating its ability to adapt to
application-specific misclassification asymmetries without degrading
discriminative performance.

\begin{figure}[!t]
    \centering
    \includegraphics[width=0.95\linewidth]{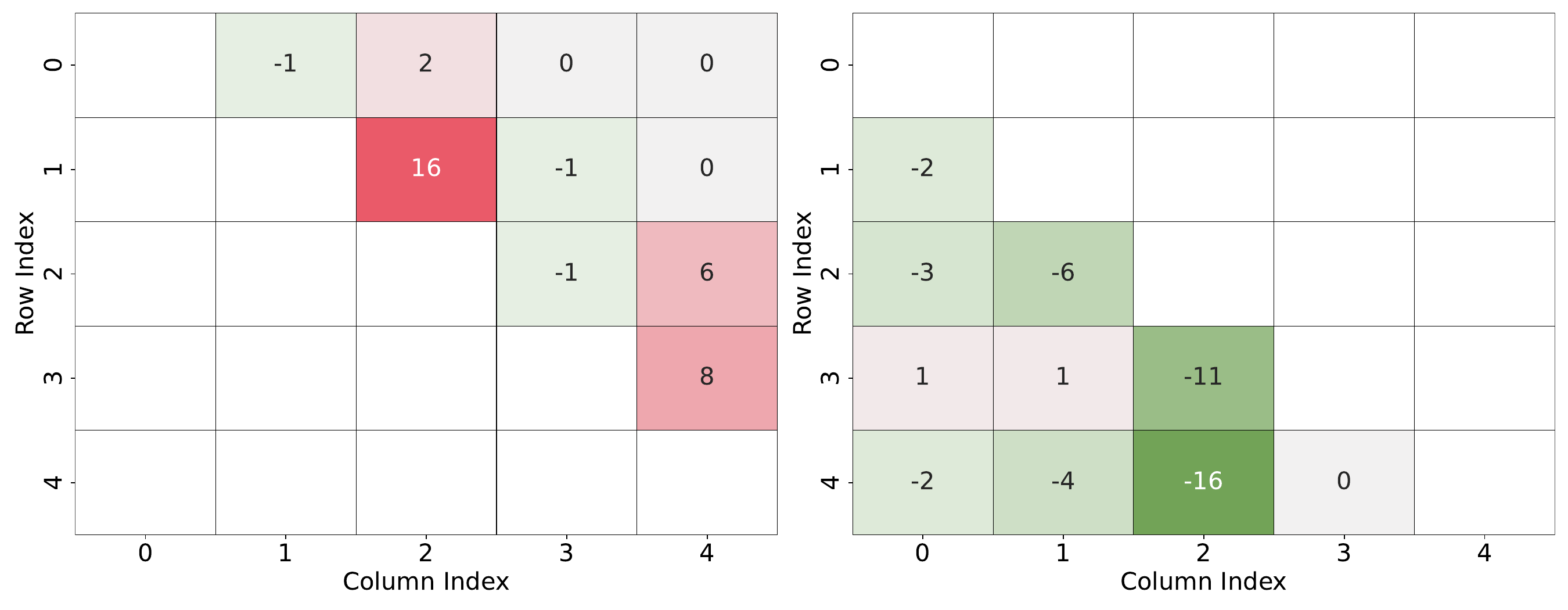}
    \caption{Impact of Asymmetric OCE on Overestimation (left) and Underestimation (right) Errors in DenseNet121}
    \label{fig:heatmap}
\end{figure}

\noindent
Figure~\ref{fig:heatmap} compares the classification behavior of the DenseNet121 model when trained with the symmetric and asymmetric variants of the OCE loss. 
The upper triangular region corresponds to overestimation errors (i.e., predicting a higher severity class than the ground truth), while the lower triangular region corresponds to underestimation errors. 
The asymmetric OCE reduces the magnitude and frequency of underestimation errors while simultaneously encouraging predictions to deviate in the correct ordinal direction. 
This behavior indicates that the asymmetric formulation imposes a directional preference that aligns model predictions more closely with the ordinal structure of the task, ultimately leading to more clinically meaningful error patterns.
The effectiveness of the proposed framework is further supported by evaluation of the impact of the diagonal term $c_{m_k,m_k}$, which represents the benefit for correct classification. When the diagonal of the penalty matrix was set to zero, the average misclassification cost increased by \SI{13.1}{\percent} in the symmetric setting and \SI{14.7}{\percent} in the asymmetric setting. This confirms that the diagonal term is essential for providing the necessary gradient signal to prioritize correct predictions alongside penalty minimization.
\section{Conclusions}
We proposed \emph{Ordinal Cross-Entropy (OCE)}, a simple and general loss function for ordinal classification that extends standard cross-entropy by incorporating distance-aware and asymmetric misclassification costs. OCE preserves the probabilistic interpretation and optimization advantages of cross-entropy while enabling explicit modeling of ordinal structure and clinically meaningful risk.

Theoretical analysis showed that OCE induces smooth and stable gradient
dynamics, encouraging confident predictions and penalizing high-risk ordinal
errors proportionally to their severity and direction. Empirical results
demonstrated that OCE consistently achieves lower misclassification cost than
state-of-the-art ordinal approaches across multiple architectures,
under symmetric and asymmetric risk settings, while maintaining competitive
performance on standard metrics.

This study has several limitations. First, the evaluation is conducted on a single medical dataset, and additional validation across diverse domains would further strengthen the generality of the findings. Second, the cost matrix parameters are predefined and not learned adaptively, which may limit flexibility in certain applications.
Future work will focus on integrating adaptive mechanisms that dynamically adjust the parameters of the OCE loss during training. Such mechanisms may allow the loss function to progressively emphasize clinically critical errors and further improve ordinal consistency and risk-aware performance.

\section*{Acknowledgment}
Gemini AI was used for language and grammar editing.

\vspace{12pt}

\clearpage
\onecolumn
\appendix[Derivation of the OCE Gradient]

In this appendix, we provide the detailed derivation of the gradient expression given in Equation~\eqref{eq:oce_grad}.

\vspace{12pt}

\begin{equation*}
\begin{aligned}
&OCE_k &&= -\left[
c_{m_k, m_k} z_{k, m_k}
- c_{m_k, m_k} \log \left( \sum_{j=1}^{I} e^{z_{k,j}} \right)
+ \sum_{\substack{i=1 \\ i \neq m_k}}^{I} c_{m_k,i} \left( 
\log \left( \sum_{j=1}^{I} e^{z_{k,j}} - e^{z_{k,i}} \right)
- \log \left( \sum_{j=1}^{I} e^{z_{k,j}} \right)
\right)
\right]. \\
\\
&\nabla_{z_{k,v}} OCE_k &&= -c_{m_k,m_k} \nabla_{z_{k,v}} z_{k,m_k}
+ \frac{c_{m_k,m_k} e^{z_{k,v}}}{\sum_{j=1}^{I} e^{z_{k,j}}}
\\
& &&-\sum_{i \ne m_k} \frac{c_{m_k,i} \nabla_{z_{k,v}} \left( \sum_{j=1}^{I} e^{z_{k,j}} - e^{z_{k,i}} \right)}{\sum_{j=1}^{I} e^{z_{k,j}} - e^{z_{k,i}}}
+ \sum_{i \ne m_k} \frac{c_{m_k,i} \nabla_{z_{k,v}} \sum_{j=1}^{I} e^{z_{k,j}}}{\sum_{j=1}^{I} e^{z_{k,j}}} \\
\\
& &&= -c_{m_k,m_k} \nabla_{z_{k,v}} z_{k,m_k}
+ \frac{c_{m_k,m_k} e^{z_{k,v}}}{\sum_{j=1}^{I} e^{z_{k,j}}}
- \sum_{i \ne m_k} \frac{c_{m_k,i} (e^{z_{k,v}} - \nabla_{z_{k,v}} e^{z_{k,i}})}{\sum_{j=1}^{I} e^{z_{k,j}} - e^{z_{k,i}}}
+ \sum_{i \ne m_k} \frac{c_{m_k,i} e^{z_{k,v}}}{\sum_{j=1}^{I} e^{z_{k,j}}} \\
\\
& &&= -c_{m_k,m_k} \nabla_{z_{k,v}} z_{k,m_k}
+ c_{m_k,m_k} \hat{p}_{k,v}
- \sum_{i \notin \{m_k, v\}} \frac{c_{m_k,i} e^{z_{k,v}}}{\sum_{j=1}^{I} e^{z_{k,j}} - e^{z_{k,i}}}
+ \hat{p}_{k,v} \sum_{i \ne m_k} c_{m_k,i} \\
\\& &&= 
 - c_{m_k,m_k} \delta_{v,m_k} 
+ c_{m_k,m_k} \hat{p}_{k,v} 
 - \hat{p}_{k,v} \sum_{i \notin \{m_k, v\}}
    \frac{c_{m_k,i}}{1 - \hat{p}_{k,i}}
+ \hat{p}_{k,v} \sum_{i \ne m_k} c_{m_k,i} \\
\\
& && \delta_{v,m_k} = 
\begin{cases}
1, & \text{if } v = m_k, \\
0, & \text{otherwise.}
\end{cases}
\end{aligned}
\end{equation*}

\newpage
\twocolumn

\end{document}